\documentclass[10pt,twocolumn,letterpaper]{article}

\usepackage{cvpr}              

\definecolor{cvprblue}{rgb}{0.21,0.49,0.74}
\usepackage[pagebackref,breaklinks,colorlinks,allcolors=cvprblue]{hyperref}
\usepackage{natbib}
\usepackage{colortbl}
\definecolor{lightgreen}{rgb}{0.84, 1.0, 0.84}
\usepackage{multirow}

\usepackage{pifont} 
\usepackage{booktabs} 
\newcommand{\cmark}{\ding{51}} 
\usepackage{adjustbox}
\usepackage{arydshln}
\usepackage{tabularx}

\usepackage{pifont} 

\usepackage{tcolorbox}
\tcbuselibrary{breakable,skins} 

\newtcbox{\hlblue}{on line, arc=1pt, colback=blue!12,  boxrule=0pt, boxsep=1pt,
  left=2pt,right=2pt, top=1pt,bottom=1pt}
\newtcbox{\hlyellow}{on line, arc=1pt, colback=yellow!12, boxrule=0pt, boxsep=1pt,
  left=2pt,right=2pt, top=1pt,bottom=1pt}
\newtcbox{\hlgreen}{on line, arc=1pt, colback=green!12, boxrule=0pt, boxsep=1pt,
  left=2pt,right=2pt, top=1pt,bottom=1pt}
\newtcbox{\hlorange}{on line, arc=1pt, colback=orange!12, boxrule=0pt, boxsep=1pt,
  left=2pt,right=2pt, top=1pt,bottom=1pt}
\newtcbox{\hlpurple}{on line, arc=1pt, colback=purple!12, boxrule=0pt, boxsep=1pt,
  left=2pt,right=2pt, top=1pt,bottom=1pt}
\newtcolorbox{promptbox}{
  breakable,                  
  width=\linewidth,
  colback=gray!10,            
  colframe=gray!30,           
  arc=2mm,                    
  boxrule=0.3pt,              
  left=6pt,right=6pt,top=6pt,bottom=6pt
}

\def\paperID{15295} 
\def\confName{CVPR}
\def\confYear{2025}

\title{HiEAG: Evidence-Augmented Generation for Out-of-Context Misinformation Detection}

\author{
Junjie Wu$^1$ \ \  \  Yumeng Fu$^3$ \ \  \    Nan Yu$^1$  \ \  \   Guohong Fu$^{1,2}$\thanks{Corresponding author.}\\
$^1$School of Computer Science and Technology, Soochow University\\
$^2$Institute of Artificial Intelligence, Soochow University\\
$^3$School of Computer Science and Technology, Harbin Institute of Technology\\
{\tt\small \{20224027010, nyu\}@stu.suda.edu.cn, 24b303004@stu.hit.edu.cn, ghfu@suda.edu.cn}
}

\begin{document}
\maketitle

\begin{abstract}
Recent advancements in multimodal out-of-context (OOC) misinformation detection have made remarkable progress in checking the consistencies between different modalities for supporting or refuting image-text pairs. However, existing OOC misinformation detection methods tend to emphasize the role of internal consistency, ignoring the significant of external consistency between image-text pairs and external evidence. In this paper, we propose HiEAG, a novel Hierarchical Evidence-Augmented Generation framework to refine external consistency checking through leveraging the extensive knowledge of multimodal large language models (MLLMs). Our approach decomposes external consistency checking into a comprehensive engine pipeline, which integrates reranking and rewriting, apart from retrieval. Evidence reranking module utilizes Automatic Evidence Selection Prompting (AESP) that acquires the relevant evidence item from the products of evidence retrieval. Subsequently, evidence rewriting module leverages Automatic Evidence Generation Prompting (AEGP) to improve task adaptation on MLLM-based OOC misinformation detectors. Furthermore, our approach enables explanation for judgment, and achieves impressive performance with instruction tuning. Experimental results on different benchmark datasets demonstrate that our proposed HiEAG surpasses previous state-of-the-art (SOTA) methods in the accuracy over all samples.
\end{abstract}    
\section{Introduction}
\label{sec:intro}

Out-of-context (OOC) misinformation on social media is particularly prevalent, and often induces significant risks, such as conspiracy theories and societal safety \cite{ma-etal-2024-event,nan2025exploiting}. This prominent type of misinformation refers to a genuine image used in a misleading or incorrect textual context, posing a unique challenge in the digital era \cite{aslett2024online,lakara2025llmconsensusmultiagentdebatevisual}. For example, malicious actors pair election images with unrelated textual context during the U.S. presidential election, thereby deceiving or misleading voters \cite{Xu_MMOOC}. To address these potential risks, it is essential to develop effective approaches for detecting OOC misinformation.

\begin{figure*}[t]
\centering
\includegraphics[width=\linewidth]{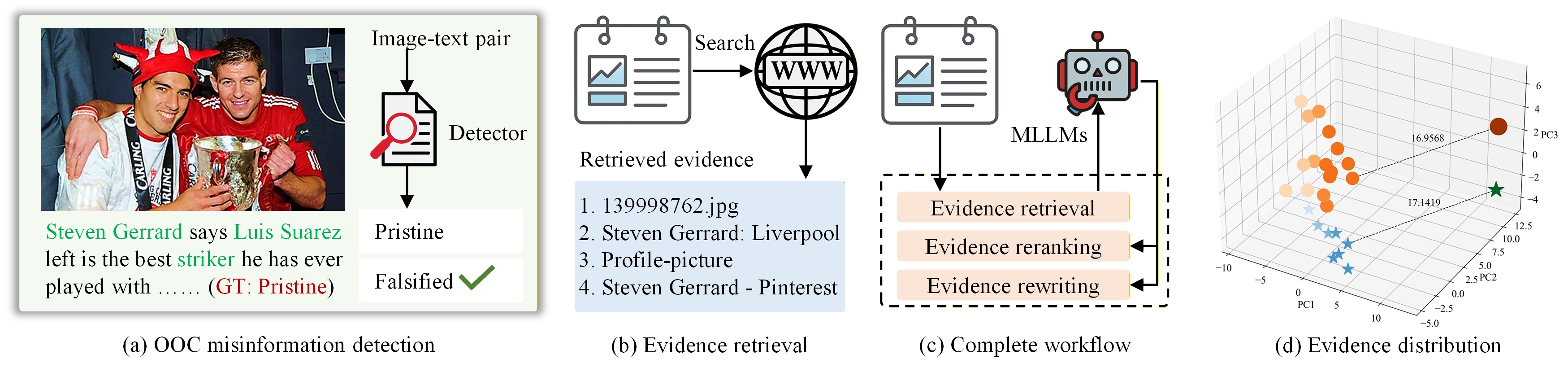}
\caption{\textbf{Multimodal OOC misinformation detection consists of critical components.} Subfigure (a) presents internal consistency checking for identifying the authenticity of an image-text pair. Subfigure (b) presents evidence retrieval through tool usage. Subfigure (c) presents core parts within the complete workflow. Subfigure (d) visualizes the Euclidean distance between image-text pairs (dark points) and their retrieved evidence (shallow points).}
\label{fig:1}
\end{figure*}

In the existing literature, the goal of OOC misinformation detection is to identify the misuse of a genuine image within a certain textual context. Early OOC misinformation detection methods primarily underscore internal consistency \cite{luo-etal-2021-newsclippings,Aneja_Bregler_Niessner_2023}, \textit{i.e}., the alignment between images and their textual context at the semantic level. Despite their impressive performance, these methods struggle to capture nuanced discrepancies, such as cross-referencing temporal and entity details. Fig. \ref{fig:1}(a) provides an example of multimodal OOC misinformation detection, in which a textual context includes ``Steven Gerrard'' and ``Luis Suarez'', and a predicted label denotes the authenticity of an image-text pair. To further enhance fact-checking, some researchers attempt to retrieve external evidence into OOC misinformation detectors with different architectures, including attached classifiers \cite{Jin_2017}, pre-trained small models \cite{pmlr-v139-radford21a}, and Multimodal Large Language Models (MLLMs) \cite{Dai_InstructBLIP}. As shown in Fig. \ref{fig:1}(b), external evidence of image-text pairs is collected through an inverse image search, and is then used to improve detection performance. Though they can work together with web-retrieved evidence, their capacity is limited due to the lack of external consistency, \textit{i.e}., the alignment between image-text pairs and external evidence. Hence, refining external consistency has become a pressing necessity in the task of OOC misinformation detection.

To reach the target above, some research leverages cosine similarity to calculate the relationships between image-text pairs and external evidence \cite{vector_Salton,Budack_TamperedNews}. On the other hand, recent advancements in MLLMs, Qi et al. \cite{Qi_2024_CVPR} introduced a two-pronged analysis for checking both internal and external consistencies via news-domain entity learning. Furthermore, inspired by the successful application of Retrieval-Augmented Generation (RAG) in large models \cite{10.1007/978-981-96-8180-8_25,10.1145/3701228}, recent studies combine this technique with lexical matching \cite{wu2025exclaimexplainablecrossmodalagentic,yan2025mitigatinggenaipoweredevidencepollution}, or association scoring \cite{xiao2025xfacta,li-etal-2025-cmie} to acquire the targeted evidence for judgment. However, they encounter the inherent limitations of external consistency checking as follows:

(a) \textbf{Incomplete workflow}: A standard workflow should include external evidence retrieval, reranking, and rewriting, as shown in Fig. \ref{fig:1}(c). The former two steps have been considered in existing studies, but the last step is unexplored. This is attributed to preference alignment from model evolution \cite{malviya-katsigiannis-2024-evidence}. MLLMs have strong understanding capacities towards evidence sentences, rather than evidence pieces.

(b) \textbf{Evidence entanglement}: External evidence with different degrees of relevance represents the relationships between image-text pairs and classification labels. Compared to weakly relevant, and even irrelevant evidence, relevant evidence is more useful for more accurate judgments. In Fig. \ref{fig:1}(d), a qualitative analysis is provided to present an urgent requirement: How to design an effective evidence-augmented generation (EAG) strategy to enhance performance in detection?

In this paper, we propose a novel Hierarchical Evidence-Augmented Generation (HiEAG) framework, designed to refine consistency checking for the task of multimodal OOC misinformation detection. Specifically, at the process of evidence reranking, we introduce Automatic Evidence Selection Prompting (AESP) to autonomously acquire the most relevant item once evidence retrieval for image-text pairs. Subsequently, we design Automatic Evidence Generation Prompting (AEGP). This utilizes large models to understand and generate an alignment sentence based on multimodal content and the selected evidence item. In addition, to enable the model for both explanation and judgment, we employ instruction-tuning on a constructed OOC misinformation dataset. Experiments on both synthetic and real-world datasets demonstrate the effectiveness and robustness of our proposed HiEAG in the realm of multimodal OOC misinformation detection.

In summary, our major contributions are three-fold:
\begin{itemize}
    \item We propose HiEAG, a novel multimodal OOC misinformation detection framework that effectively refines consistency checking between multimodal content and external evidence, significantly improving detection performance.
    \item We develop the AESP to leverage the knowledge of MLLM's parameters, selecting the relevant item from retrieved evidence. Additionally, the AEGP is designed to achieve the alignment sentence for task adaptation on MLLM-based OOC misinformation detection models.
    \item We conduct extensive experiments on the NewsCLIPpings and VERITE benchmark datasets, demonstrating that HiEAG outperforms the SOTA methods in the accuracy over all samples.
\end{itemize}

The rest of this paper is organized as follows. Section 2 summarizes the related works of OOC misinformation detection. Section 3 presents the details of our proposed approach HiEAG. Section 4 provides the experimental settings and results. Finally, we provide a conclusion and future research directions in Section 5.
\section{Related Work}
\label{sec:Related Work}

\subsection{OOC misinformation detection}

Out-of-context misinformation (OOC) detection is the task of identifying the misuse of a genuine image within a certain textual context. Due to the rapid expansion of forgery content on social media, this detection task is urgent to alleviate the potential risk. Recently, a large amount of OOC misinformation detection methods have been developed. The methods are categorized as: internal checking \cite{luo-etal-2021-newsclippings,Papadopoulos_DT,Aneja_Bregler_Niessner_2023} and external checking \cite{Jaiswal_MAIM,Sabir_MEIR,Jaiswal_2019_CVPR,Budack_TamperedNews,zlatkova-etal-2019-fact,Abdelnabi_2022_CVPR,yuan-etal-2023-support}. For the internal checking that focuses on modality features for capturing the semantic differences between multiple modalities, Papadopoulos et al. \cite{Papadopoulos_DT} used CLIP \cite{pmlr-v139-radford21a} and auxiliary Transformer layers to intensify multimodality feature extraction and interaction. Aneja et al. \cite{Aneja_Bregler_Niessner_2023} utilized visual grounding with textual descriptions for interpreting semantic conflicts. For the external checking that introduces web-retrieved evidence into detectors for cross-modal consistency reasoning, Abdelnabi et al. \cite{Abdelnabi_2022_CVPR} established a CNN-based model to perform the cycle consistency between image-text pairs and web-retrieved evidence. Yuan et al. \cite{yuan-etal-2023-support} followed this work to further learn the stances of external evidence through neural networks.

Despite remarkable progress made in previous OOC misinformation detection methods, the accuracy of detection is still limited by the scale of pre-trained multimodal models.

\subsection{MLLM-based OOC misinformation detection}

Against the backdrop of the rapid development of Large Language Models (LLMs), MLLMs have emerged as a promising alternative for visual and multimodal tasks \cite{Jiang_2024_CVPR, Liu_2024_CVPR, Liu_UFAFormer}. MLLMs likewise LLaVA \cite{Liu_LLaVA}, MiniGPT-v2 \cite{chen2023minigptv2largelanguagemodel}, and LLaVA-1.5 \cite{Liu_LLaVA15}, have demonstrated remarkable multimodal understanding capabilities, and presented superior zero-shot performance in various tasks, such as Visual Question Answer (VQA) \cite{Han_ZSVQA, YUAN2025111783} and Anomaly Segmentation (AS) \cite{LI2025129122, PENG2025113176}. For instance, Xu et al. \cite{Xu_ZSVQA} designed prompts to extract visual information for answering visual questions. Xu et al. \cite{Xu_Sun_Zhai_Li_Liang_Li_Du_2025} introduced a tuning-free pipeline to leverage frozen MLLMs for video moment retrieval. Some studies \cite{zhou2024correctingmisinformationsocialmedia, li-etal-2025-cmie, wu2025exclaimexplainablecrossmodalagentic, gu2025multimllmknowledgedistillationoutofcontext} leverage the closed-source MLLMs and expensive API calls to detect OOC misinformation in the zero-shot or few-shot setting. Such manner is impractical and inefficient in low-resource scenarios. A Parameter-Efficient Fine-Tuning (PEFT) technique, namely instruction tuning, is introduced for MLLMs on task-specific adaptation. Qi et al. \cite{Qi_2024_CVPR} introduced InstructBLIP \cite{Dai_InstructBLIP} with Vicuna \cite{vicuna2023} to conduct both internal and external consistency checking. Inspired by this, Xuan et al. \cite{xuan2024lemmalvlmenhancedmultimodalmisinformation} adopted multi-query generation to gather comprehensive evidence for enhancing the detection capabilities of MLLMs. Chat-OOC \cite{10.1007/978-3-031-57916-5_8} based on MiniGPT-4 \cite{zhu2023minigpt4enhancingvisionlanguageunderstanding} with 13B parameters employs instruction tuning to improve the controllability of MLLMs for both judgments and explanations. In this work, our proposal considers the extensive knowledge of MLLMs to refine consistency checking between image-text pairs and external evidence, thus effectively debunking multimodal OOC misinformation.
\section{Methodology}
\label{sec:Methodology}

\subsection{Problem setting \& overview}

Given an out-of-context misinformation dataset $\mathcal{D}=\{\textbf{x}_i, y_i\}_i^N$, where $\textbf{x}_i=\{x_i^I,x_i^T\}$ denotes a news image and its attached textual context. Each pair of image-text is assigned with the ground-truth label $y_i\in\{0,1\}$, being either Pristine (Not out-of-context) or Falsified (Out-of-context). $N$ denotes the size of $\mathcal{D}$. The goal of OOC misinformation detection is to train a detector $f(\textbf{x}_i)\to\hat{y}_i$, which provides a judgment $\hat{y}_i$ regarding the authenticity of an image-text pair $\textbf{x}_i$. 

\begin{figure*}[t]
\centering
\includegraphics[width=\linewidth]{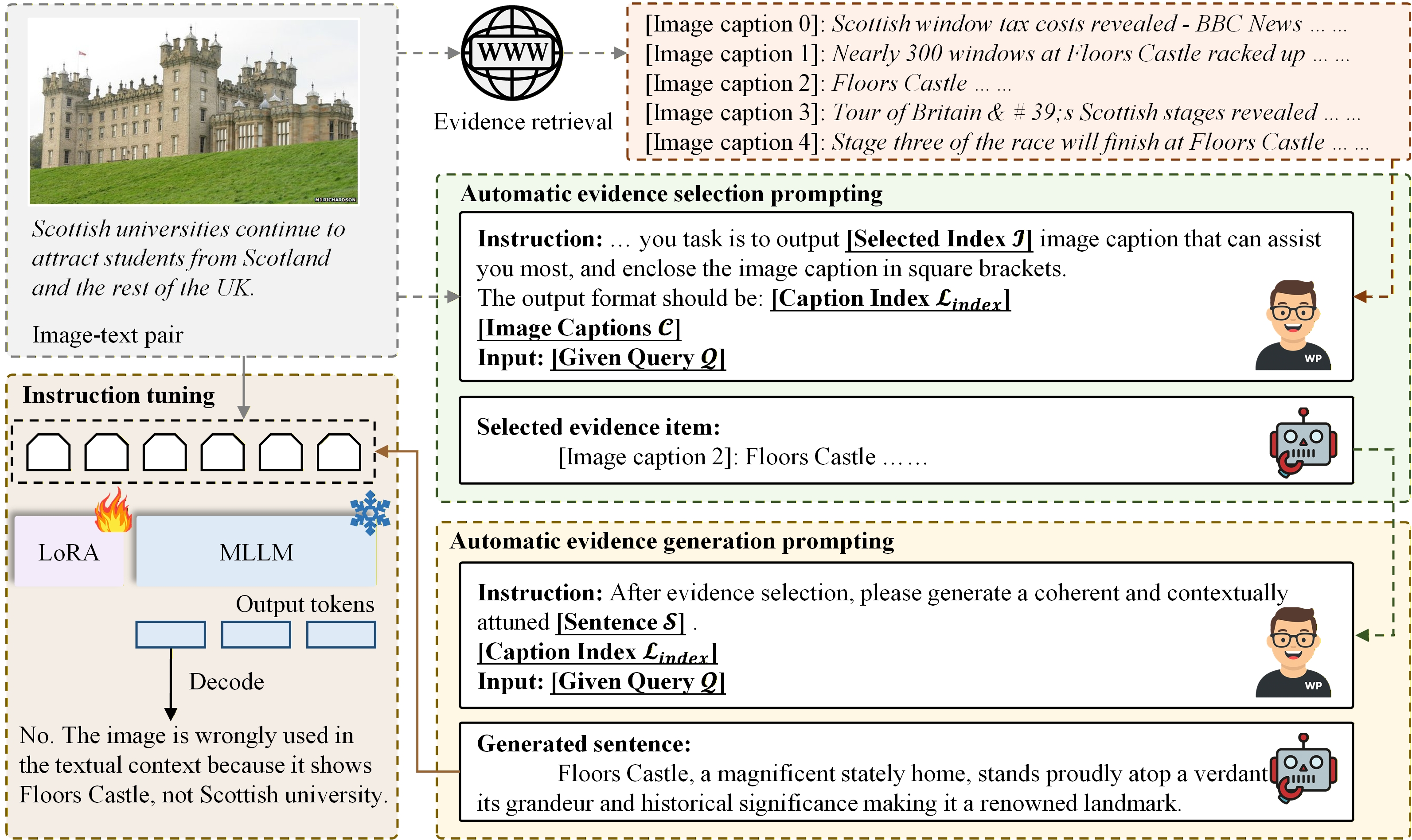}
\caption{\textbf{The overview of our proposed framework HiEAG.} (a) Automatic evidence selection prompting acquires the most relevant evidence item regarding the image-text pair. (b) Automatic evidence generation prompting achieves a novel sentence that aligns the selected item. (c) Instruction tuning enables the MLLM-based OOC misinformation detector for both judgment and explanation.}
\label{fig:2}
\end{figure*}

As depicted in Fig. \ref{fig:2}, we present a novel hierarchical framework HiEAG for OOC misinformation detection. HiEAG begins with evidence retrieval that stems from the Internet. External evidence is then processed through rerank-and-rewrite prompting. In the Automatic Evidence Selection Prompting (AESP), the most relevant evidence item is capturing by query reranking, rather than calculating the Euclidean distance between multimodality vectors. In the Automatic Evidence Generation Prompting (AEGP), the generated sentence aligns the original meaning of the selected evidence, thereby assisting in checking the consistencies between image-text pairs and external evidence. The two prompting mechanisms jointly contribute to the final judgment. The former enhances evidence retrieval while the latter highlights evidence optimization for task adaptation on large models. This designed hierarchical decomposition refines consistency checking between multimodal content and external information while optimizing the utilization of external evidence for debunking OOC misinformation.

\subsection{Evidence retrieval}

Evidence retrieval (ER) refers to knowledge around an image-text pair $\textbf{x}$ by querying external knowledge bases, such as the Google Vision API. The retrieved evidence is regarded as potential clues for consistency checking. In practice, we follow the previous collection pipeline \cite{Abdelnabi_2022_CVPR}, using the API to perform evidence retrieval for image captions. The API returns the containing page's URLs associated with the image. A web crawler is designed to visit these pages, and then saves the captions if found. The captions may contain the image's content and further contain the image's context. However, in rare cases where the captions is not available, we employ the MLLM to directly describe the content of the image for subsequent consistency checking.

\subsection{Automatic evidence selection prompting}

After evidence retrieval for image captions, as shown in Fig. \ref{fig:2}, we design Automatic Evidence Selection Prompting (AESP) to choose the most relevant evidence item and further leverage the MLLM's inherent knowledge. In practice, the prompt is as follows:

\begin{promptbox}
\textbf{Instruction:} \dots your task is to output \hlyellow{[Selected Index $\mathcal{I}$]} image caption that can assist you most, and enclose the image caption in square brackets.\par
The output format should be: \hlgreen{[Caption Index $\mathcal{L}_{\textit{index}}$]}

\hlblue{[Image Captions $\mathcal{C}$]}

\medskip
\textbf{Input:} \; \hlorange{[Given Query $\mathcal{Q}$]}
\end{promptbox}

In practice, the AESP directly inputs the \hlorange{[Given Query $\mathcal{Q}$]} and the set of \hlblue{[Image Captions $\mathcal{C}$]} into the MLLM for analyzing the relationships between image-text pairs and external information. After that, the \hlyellow{[Selected Index $\mathcal{I}$]} of image captions is integrated into the final output \hlgreen{[Caption Index $\mathcal{L}_{\textit{index}}$]}. Formally, the evidence selection process is presented as follows:
\begin{equation}
\mathcal{L}_{\textit{index}} = \operatorname*{argmax}_{\mathcal{I}\in\mathcal{C}_{\textit{range}}}\ p(\mathcal{I} \mid \mathcal{Q}, \mathcal{C}),
\end{equation}
where $\mathcal{C}_{\textit{range}}$ represents the number of image captions, and $\mathcal{L}_{\textit{index}}$ represents the selected item to facilitate the final decision-making. 

\subsection{Automatic evidence generation prompting}

Once achieving the most relevant evidence item through the AESP, as shown in Fig. \ref{fig:2}, we further introduce Automatic Evidence Generation Prompting (AEGP). In practice, our carefully devised prompt for guiding the MLLM to automatically generate a novel sentence while retaining the original meaning of the selected evidence item as follows:

\begin{promptbox}
\textbf{Instruction:} After evidence selection, please generate a coherent and contextually attuned \hlpurple{[Sentence $\mathcal{S}$]}.

\hlgreen{[Caption Index $\mathcal{L}_{\textit{index}}$]}

\medskip
\textbf{Input:} \; \hlorange{[Given Query $\mathcal{Q}$]}
\end{promptbox}

In practice, the process depends on the \hlorange{[Given Query $\mathcal{Q}$]} and the \hlgreen{[Caption Index $\mathcal{L}_{\textit{index}}$]} to directly guide the MLLM to generate a coherent and contextually attuned \hlpurple{[Sentence $\mathcal{S}$]}. This enhances the performance of detectors by aligning the image caption to the evidence sentence generated from the MLLM. Formally, the evidence generation process is presented as follows:
\begin{equation}
\mathcal{S} = \operatorname*{argmax}\ p(s \mid \mathcal{Q}, \mathcal{L}_{\textit{index}}),
\end{equation}
where $\mathcal{S}$ represents the alignment sentence for the selected item of the AESP. The proposed hierarchical framework consists of both AESP and AEGP, which contributes to evidence-augmented generation for refining consistency checking. Ablation study demonstrates that utilizing this proposal can effectively improve detection performance.

\subsection{Evidence-augmented instruction dataset}

\begin{figure}[t]
    \centering
    \includegraphics[width=\linewidth]{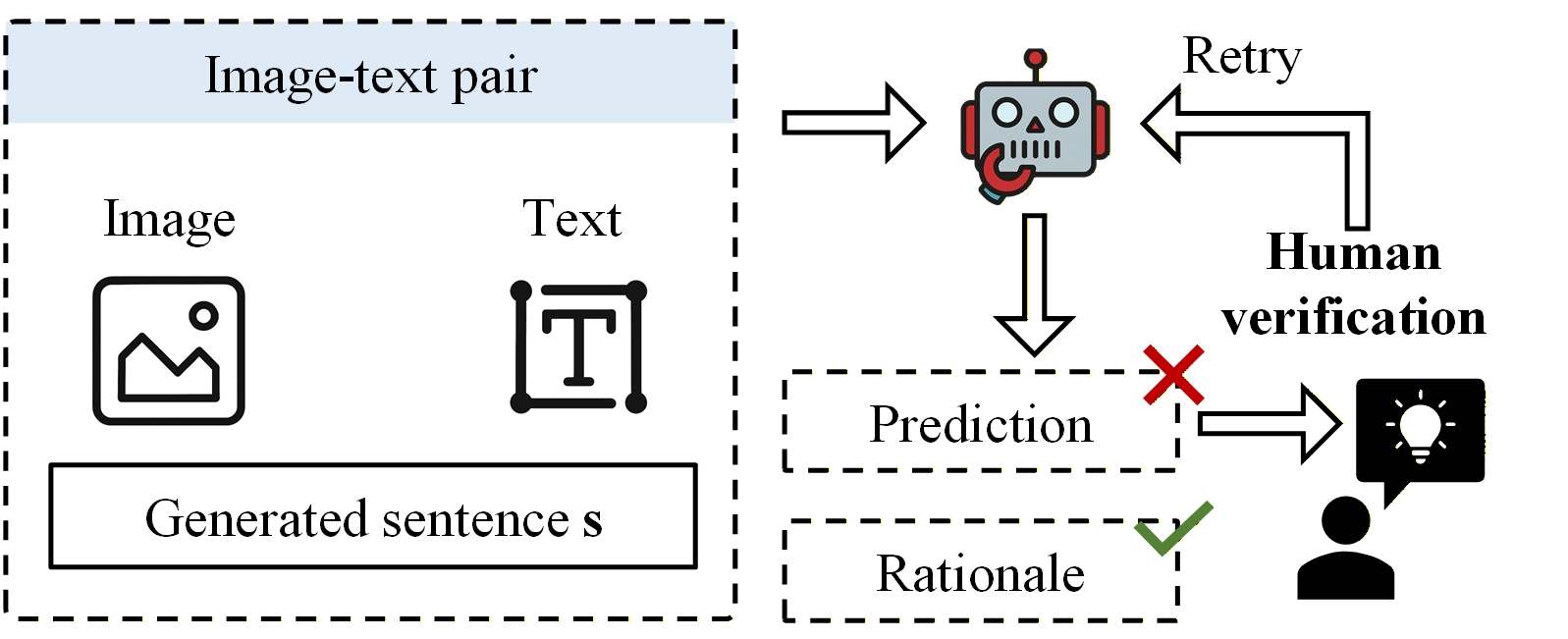}
    \caption{The construction process of evidence-augmented instruction dataset.}
\label{fig:3}
\end{figure}

To enable the explanation capabilities of detectors for more accurate judgments, we provide a detailed process of constructing the instruction dataset for multimodal OOC misinformation detection, as shown in Fig. \ref{fig:3}. The instruction dataset is built upon the NewsCLIPpings dataset \cite{luo-etal-2021-newsclippings}. In practice, we present two categories of targeted outputs $y'$. For each pristine sample, we directly provide the targeted language without any rationales (\textit{i.e}., ``Yes. The image is rightly used in the textual context.'' ). For each falsified sample, we achieve the predicted language with a rationale (\textit{i.e}., ``No. The image is wrongly used in the textual context. $\underline{\textit{rationale}}$'' ). Formally, the predicted language can be calculated as follows:
\begin{equation}
y' = \text{MLLMs}(\textbf{x}, \textbf{s}, q),
\label{Eq:3}
\end{equation}
where $\textbf{x}$ represents the image-text pair, $\textbf{s}$ represents the alignment sentence, and $q$ represents the given query. After achieving the predicted output, we further perform a verification process between the predicted output $y'$ and the correct label $y$. This is to ensure that the predicted output matches the correct label. If there are inconsistencies between them, we retry Eq. (\ref{Eq:3}) until they are equal. Finally, the constructed instruction dataset $\mathcal{D}'=\{\textbf{x}_i, y'_i\}_i^N$ consists of 35,536 pristine samples and 35,536 falsified samples with rationales.

\subsection{Evidence-augmented instruction tuning}

After constructing the instruction dataset, we further adopt instruction tuning with the LoRA adapter \cite{hu2022lora} to the task of multimodal OOC misinformation detection. This preserves the original parameters $\theta$ of the model, thereby reducing the computational resource usage:
\begin{equation}
\theta' = \theta + \Delta\theta, \ \ \Delta\theta = \text{LoRA}(A, B),
\end{equation}
where $\Delta\theta$ represents the updated model parameters, $A\in \mathbb{R}^{r\times d}$ and $B\in \mathbb{R}^{d\times r}$ represent low rank matrices. The signs $d$ and $r$ denote the dimension and the rank within a single matrix, respectively.

In accordance with the optimization objective of MLLMs, we apply the next token prediction loss to assess the output error of the task-specific MLLM, which can be formulated as follows:
\begin{equation}
\mathcal{L} = -\frac{1}{B} \frac{1}{T} \sum_{i=1}^{B} \sum_{t=1}^{T} \text{log}\ P(\epsilon_t \mid \epsilon_{<t}, x'),
\end{equation}
where $B$ and $T$ represent the batch size and the number of tokens, respectively. The sign $x'$ represents the input sequence that integrates both textual and visual representations. This serves as the context for next token predictions.
\section{Experiment}
\label{sec:Experiment}

In this section, we conduct extensive experiments to evaluate the effectiveness of our proposed HiEAG in the task of OOC misinformation detection. First, we briefly introduce benchmark datasets and evaluation metrics used in the experiments, followed by a discussion on the compared methods and training details.

\subsection{Experimental setup}

\textbf{Datasets.} We conduct experiments on two widely used OOC misinformation datasets, including NewsCLIPpings \cite{luo-etal-2021-newsclippings} and VERITE \cite{papadopoulos2024verite}.
\begin{itemize}[label=\Large\textbullet]
    \item \textbf{NewsCLIPpings} \cite{luo-etal-2021-newsclippings} is the largest benchmark dataset for OOC misinformation detection. It is derived from the dataset VisualNews \cite{liu-etal-2021-visual}, which stems from four news agencies, \textit{i.e.,} The Guardian, BBC, The Washington Post and USA Today. The out-of-context samples in this dataset are generated by replacing original images with semantically related images but present distinct news events. The NewsCLIPpings Merged-Balanced subset is divided into 71,072 samples for training, 7,024 samples for validation, and the remaining 7,264 samples for testing, respectively. Each sample contains a pair of image-text, and the dataset is evenly balanced with respect to class labels.
    \item \textbf{VERITE} \cite{papadopoulos2024verite} is a real-world dataset, which consists of 1,000 annotated samples in the form of image-text pairs. These samples are collected from fact-checking websites, such as Snopes and Reuters. In these websites, human experts verify the authenticity of content in cycle. Each image is paired with different textual context. One is regarded as pristine, and other is verified as falsified. Moreover, to avoid uni-modal bias, this dataset excludes asymmetric multimodal content, and adopts modality balancing strategy. To align with previous works in experimental settings, we also train the proposed HiEAG on the training set of NewsCLIPpings, and then report its results on the NewsCLIPpings test set and VERITE, respectively.
\end{itemize}

\noindent\textbf{Evaluation metrics.} To evaluate the performance of our proposed method HiEAG in the task of OOC misinformation detection, we utilize accuracy across all samples (All), separately for the OOC (Falsified) samples and the not OOC (Pristine) samples as performance metrics. These metrics are regarded as a comprehensive assessment of detectors efficiency \cite{luo-etal-2021-newsclippings}. The accuracy metric denotes the proportion of correctly classified instances. In judgment, a professional OOC misinformation detector should maintain a good balance between Falsified samples and Pristine samples.

\noindent\textbf{Baseline methods.} To execute a comprehensive evaluation of our proposed method HiEAG, we compare it with several representative OOC misinformation detection methods. The compared methods can be categorized as follows:

\noindent\textit{Auxiliary learning methods.} This category contains \textbf{SAFE} \cite{10.1007/978-3-030-47436-2_27} and \textbf{EANN} \cite{10.1145/3219819.3219903}. The former method designs an auxiliary loss to calculate sentence similarity, and translates images into descriptive sentences. The latter method adopts an adversarial learning framework to combine the event discrimination loss with the detection loss. These methods leverage auxiliary tasks to effectively train detectors for identifying multimodal OOC misinformation.

\noindent\textit{Pre-trained methods.} This category includes \textbf{VisualBERT} \cite{li2019visualbertsimpleperformantbaseline}, \textbf{CLIP} \cite{pmlr-v139-radford21a}, \textbf{Neu-Sym Detector} \cite{zhang2024interpretabledetectionoutofcontextmisinformation}, \textbf{DT-Transformer} \cite{Papadopoulos_DT}, \textbf{CCN} \cite{Abdelnabi_2022_CVPR}, and \textbf{SEN} \cite{yuan-etal-2023-support}. VisualBERT leverages a unified transformer to optimize the alignment of multimodal content. CLIP adopts contrastive learning to acquire similar representations of image-text pairs. Neu-Sym Detector performs neural-symbolic reasoning through text decomposition and aggregation. DT-Transformer introduces more transformer layers to refine the interaction of multimodal content. CCN based on CLIP conducts multimodal cycle-consistency check. SEN models the stance extraction and calculates support-refutation scores for judgments.

\noindent\textit{MLLM-based methods.} This group encompasses \textbf{Chat-OOC} \cite{10.1007/978-3-031-57916-5_8} and \textbf{SNIFFER} \cite{Qi_2024_CVPR}. Chat-OOC leverages a cross entropy loss to fine-tune a MLLM in the realm of OOC misinformation detection. SNIFFER adopts a two-stage instruction tuning, and directly integrates retrieved evidence for the final judgment. The methods rely on MLLM's promising capabilities for detecting OOC misinformation.

\noindent\textbf{Implementation details.} We employ PandaGPT (7B) \cite{su-etal-2023-pandagpt} and Qwen2-VL (7B) \cite{Qwen2VL} models as the base MLLMs. We implement HiEAG on PyTorch \cite{10.5555/3454287.3455008} version 2.3.1 with CUDA 12.2, and conduct all experiments with 4 NVIDIA GeForce RTX 3090 (24G) GPUs. The model is optimized by AdamW \cite{loshchilov2019decoupledweightdecayregularization} with a liner warm-up learning rate strategy. We set the batch size to 8, the learning rate to 2e-4, and train the model for 5 epochs. In addition, we leverage FlashAttention-2 \cite{dao2023flashattention2} to replace the original attention layers of large models for efficient training.

\subsection{Results on NewsCLIPpings dataset}

Table \ref{tab:1} provides performance comparison between HiEAG and other methods on the NewsCLIPpings \cite{luo-etal-2021-newsclippings} test set.  From the results, we can obtain the following observations as: (1) HiEAG achieves superior performance, which outperforms other methods in terms of all samples. (2) HiEAG with different base MLLMs consistently exhibits significant gains in all evaluation metrics. (3) HiEAG employs 7B parameters of large models, which is less than that of other MLLM-based methods, such as Chat-OOC and SNIFFER. (4) HiEAG presents a good balance between falsified samples and pristine samples. Overall, our proposed method HiEAG has strong detection accuracy in OOC misinformation, achieving performance significantly better than the base MLLMs. This demonstrates the efficacy of HiEAG.

\begin{table}[t]
\centering
\caption{\textbf{Comparison of OOC misinformation detection methods on the test set of NewsCLIPpings \cite{luo-etal-2021-newsclippings}.} For each metric, we report the average of three runs without any hyperparameter searching. The best results are indicated in \textbf{bold}.}
\begin{adjustbox}{valign=c, max width=\columnwidth}
\begin{tabular}{l|cccc}
\toprule
Method  & \textbf{All $\uparrow$} & \textbf{Falsified $\uparrow$} & \textbf{Pristine $\uparrow$}  \\
\hline
SAFE \cite{10.1007/978-3-030-47436-2_27}    & 52.8 & 54.8 & 52.0  \\
EANN \cite{10.1145/3219819.3219903}   & 58.1 & 61.8 & 56.2  \\
VisualBERT \cite{li2019visualbertsimpleperformantbaseline}   & 58.6 & 38.9 & 78.4 \\
CLIP \cite{pmlr-v139-radford21a}  & 66.0 & 64.3 & 67.7  \\
Neu-Sym detector \cite{zhang2024interpretabledetectionoutofcontextmisinformation}  & 68.2 & - & - \\
DT-Transformer \cite{Papadopoulos_DT}  & 77.1 & 78.6 & 75.6  \\
CCN \cite{Abdelnabi_2022_CVPR}   & 84.7 & 84.8 & 84.5  \\
SEN \cite{yuan-etal-2023-support}  & 87.1 & 85.5 & 88.6  \\
Chat-OOC (13B) \cite{10.1007/978-3-031-57916-5_8} & 80.0 & - & - \\
InstructBLIP (13B) \cite{Dai_InstructBLIP}   & 82.5 & 75.3 &	89.7 \\
SNIFFER \cite{Qi_2024_CVPR}   & 88.4 & 86.9 & \textbf{91.8}  \\
\hline
PandaGPT (7B) \cite{su-etal-2023-pandagpt}   & 72.7 & 72.1 & 73.2 \\
 \ \ w/\ HiEAG   & 84.2 & 84.0 & 84.4 \\
\hline
Qwen2-VL (7B) \cite{Qwen2VL}  & 78.9 & 73.8 & 84.2 \\
\ \ w/\ HiEAG  & \textbf{89.9} & \textbf{90.3} & 89.4  \\
\bottomrule
\end{tabular}
\end{adjustbox}
\label{tab:1}
\end{table}

\begin{table}[t]
\caption{\textbf{Comparison of OOC misinformation detection methods on the test set of VERITE \cite{papadopoulos2024verite}.} T/O denotes true versus out-of-context. For each metric, we report the average of three runs without any hyperparameter searching. The best results are indicated in \textbf{bold}.}
      \centering
      \begin{tabular}{l|c}
        \toprule
        Method  & \textbf{T/O $\uparrow$}\\
        \hline
        VERITE \cite{papadopoulos2024verite} & 72.7 \\
        SNIFFER (InstructBLIP 13B) \cite{Qi_2024_CVPR}   &  74.0 \\
        \hline
        HiEAG (Qwen2-VL 7B)   & \textbf{74.4} \\
        \bottomrule
  \end{tabular}
  \label{tab:2}
\end{table}

\subsection{Results on VERITE dataset}

Furthermore, we compare our HiEAG method with other methods such as VERITE \cite{papadopoulos2024verite} and SNIFFER \cite{Qi_2024_CVPR} on a real-world benchmark dataset VERITE \cite{papadopoulos2024verite}. As shown in Table \ref{tab:2}, HiEAG outperforms all methods. We attribute this to superior evidence-augmented generation approach in the realm of OOC misinformation detection. Typically, compared to the method VERITE, SNIFFER and HiEAG underscore the significance of evidence retrieval related to multimodal content. In evidence entanglement, our method acquires the relevant evidence, rather than all evidence, thereby refining consistency checking between multimodal content and external information. This superior performance indicates the effectiveness and generalizability capabilities of our proposed method HiEAG, even in a real-world OOC misinformation dataset.

\subsection{Ablation study}

\begin{table}[t]
\centering
\caption{\textbf{Ablation experiments for the HiEAG with the base MLLM PandaGPT (7B) on the NewsCLIPpings \cite{luo-etal-2021-newsclippings} test set.} ER and RE denote evidence retrieval and reasoning explanation, respectively.}
\begin{adjustbox}{valign=c, max width=\columnwidth}
\setlength{\tabcolsep}{2.1pt} 
        \begin{tabular}{ccccc|ccc}
            \toprule
            ER & AESP & AEGP & RE & Tuning & \textbf{All $\uparrow$} & \textbf{Falsified $\uparrow$} & \textbf{Pristine $\uparrow$} \\
            \hline
              &  &  &  & & 49.4 &	56.1 & 42.8 \\
            \cmark &  &  &  & & 51.1 &	57.2 & 45.0 \\
            &  &  &  &\cmark & 72.7 &	72.1 & 73.2 \\
            \cmark &  &  &  &\cmark & 79.9 &	78.9 & 80.9 \\
            \cmark & \cmark &  &  &\cmark & 82.1 &	79.3 & 85.2 \\
            \cmark & \cmark & \cmark &  &\cmark & 83.4 &	80.8 & 86.0 \\
            \cmark & \cmark & \cmark &  \cmark &\cmark & 84.2 & 84.0 & 84.4 \\
            \bottomrule
        \end{tabular}
\end{adjustbox}
\label{tab:3}
\end{table}

\begin{table}[t]
\centering
\caption{\textbf{Ablation experiments for the HiEAG with the base MLLM Qwen2-VL (7B) on the NewsCLIPpings \cite{luo-etal-2021-newsclippings} test set.} ER and RE denote evidence retrieval and reasoning explanation, respectively.}
\begin{adjustbox}{valign=c, max width=\columnwidth}
\setlength{\tabcolsep}{2.1pt} 
        \begin{tabular}{ccccc|ccc}
            \toprule
            ER & AESP & AEGP & RE & Tuning & \textbf{All $\uparrow$} & \textbf{Falsified $\uparrow$} & \textbf{Pristine $\uparrow$} \\
            \hline
            &  &  &  & & 69.1 & 54.4 & 83.9 \\
           \cmark &  &  &  & & 76.7 &	68.0 &	85.3 \\
             &  &  &  &\cmark   & 78.9 & 73.8 & 84.2 \\
          \cmark &  &  &  &\cmark   & 83.0 & 77.1 & 88.9 \\
           \cmark & \cmark &  &  &\cmark   & 87.7 & 86.5 & 88.7  \\
           \cmark & \cmark & \cmark &  &\cmark   & 88.5 & 87.7 & 89.1  \\
           \cmark & \cmark & \cmark &  \cmark &\cmark  & 89.9 & 90.3 & 89.4 \\
            \bottomrule
        \end{tabular}
\end{adjustbox}
\label{tab:4}
\end{table}

In this study, we provide an ablation experiment to evaluate the effectiveness of critical components in our proposed method HiEAG. The experiments are conducted on the test set of NewsCLIPpings dataset, and the results are reported in Table \ref{tab:3} and Table \ref{tab:4}.

First, the experiment that presents the comparison between w ER and w/o ER in a zero-shot setting (see 1-th row and 2-th row in both Tables). Introducing evidence retrieval into the base MLLMs achieves significant gains in performance. This indicates that external evidence is potential to assist advanced detectors for detecting OOC misinformation, even in zero-shot scenarios. Second, in instruction tuning, the experiment that removes the proposed HiEAG approach demonstrates a decline in all evaluation metrics (see 3-th row to 6-th row in both Tables). This approach plays a critical role in extracting and modeling the relationships between multimodal content. Finally, the utilization of reasoning explanation also results in a gain in detection performance (see 7-th row in both Tables). This enhances the detector's capacity to provide possible explanations for judgments.

The results above demonstrate that each component in the proposed HiEAG framework is complementary. Together, the HiEAG achieves outstanding performance in multimodal OOC misinformation detection. This confirms the effectiveness of each component in learning the relevant evidence item for supporting or refuting multimodal content.

\subsection{Further analysis}

\begin{figure}[t]
    \centering
    \includegraphics[width=0.95\linewidth]{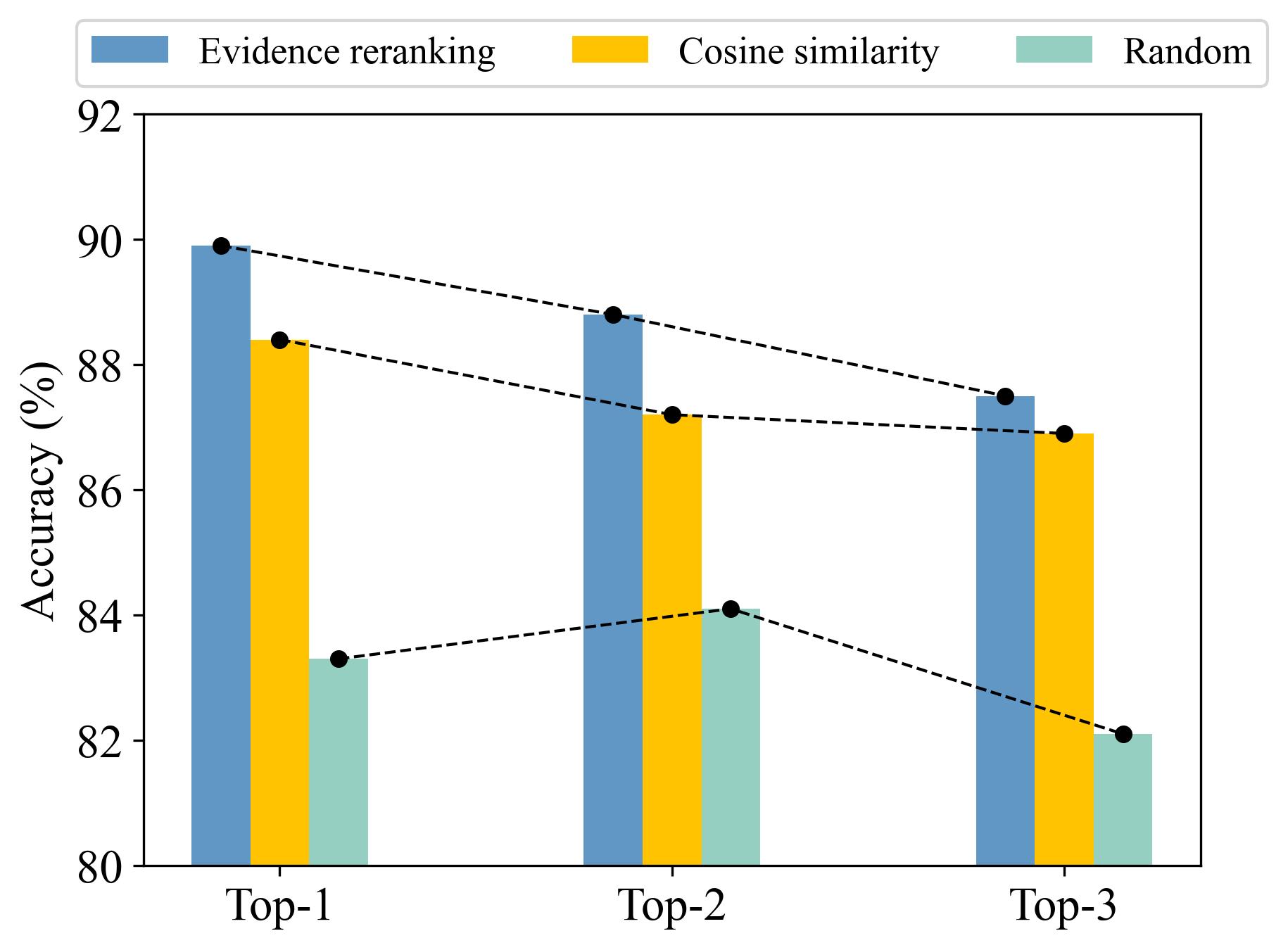}
    \caption{\textbf{Ablation experiments for HiEAG with distinct evidence reranking strategies across the NewsCLIPpings \cite{luo-etal-2021-newsclippings} dataset.} The axis denoting the number (Top-$k$) of evidence increases gradually, whereas the axis for the accuracy (\%) across all samples declines.}
\label{fig:4}
\end{figure}

\begin{figure}[t]
  \centering
   \includegraphics[width=.9\linewidth]{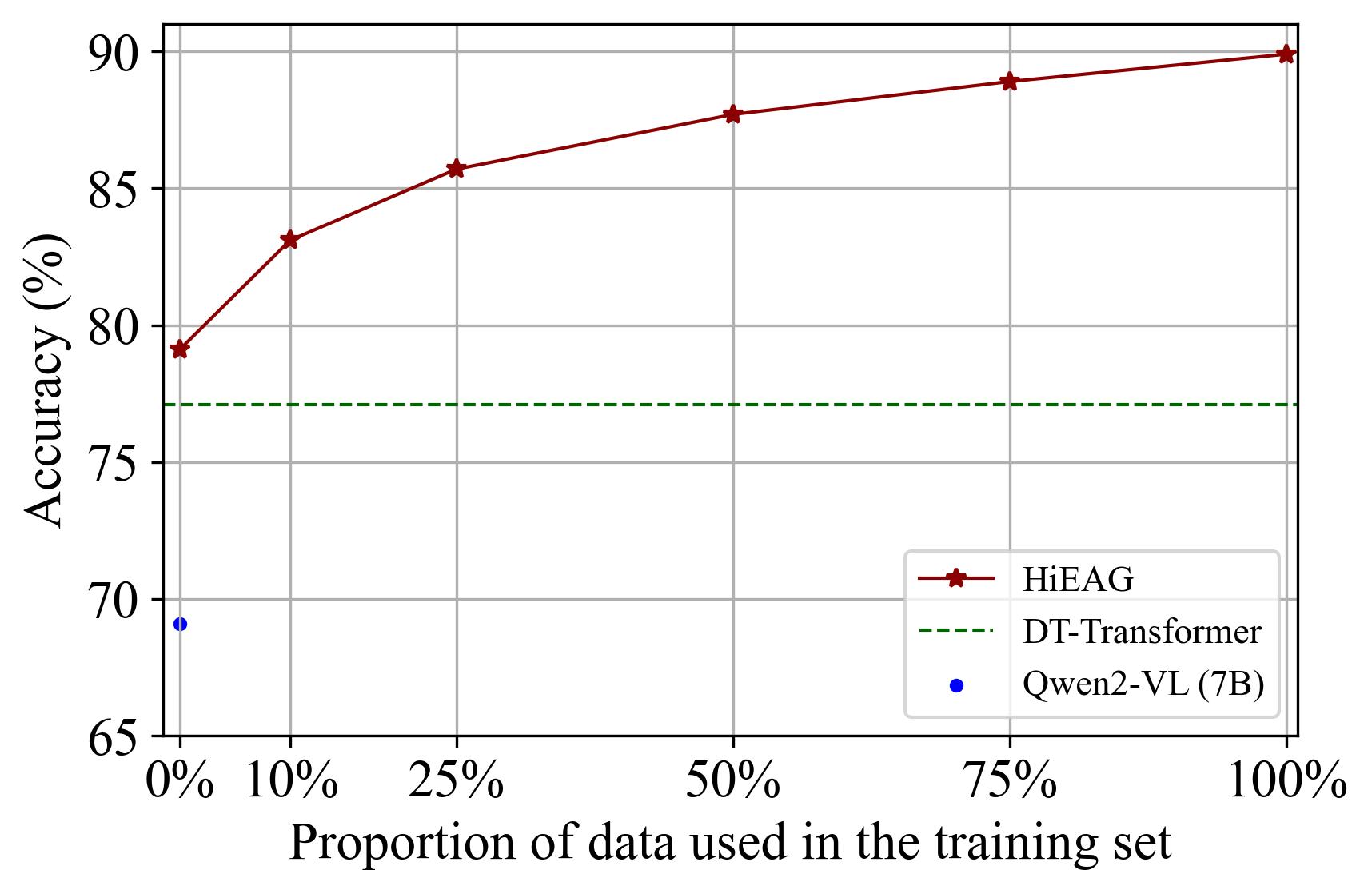}
   \caption{Performance of HiEAG on the NewsCLIPpings \cite{luo-etal-2021-newsclippings} using different training data proportions.}
   \label{fig:5}
\end{figure}

In this section, we provide an in-depth analysis of the proposed HiEAG from the perspectives of evidence reranking strategy, model size and training data size, as shown in \cref{fig:4}, \cref{fig:5}, and Table \ref{tab:5}, respectively. This is to discuss the efficiency of HiEAG.

\noindent\textbf{Analysis of evidence reranking.} We further investigate the role of evidence reranking in our proposed method HiEAG, as shown in \cref{fig:4}. In this investigation, we vary the number of external evidence from 1 to 3. This avoids the degradation of performance due to the increasement of context input. Specifically, the proposed evidence reranking strategy surpasses others consisting of random and cosine similarity across different settings. This achieves peak performance with top-1 evidence item in the term of accuracy metric. More external items exhibit limited capacity to conduct consistency checking between multimodal content and supplementary information, resulting in suboptimal performance. Compared with random and cosine similarity, our approach leverages the extensive knowledge of MLLMs parameters to acquire the relevant evidence item, thereby effectively reducing the disturbance of weakly relevant items and even irrelevant items. This suggests that the MLLM reranks external evidence based on the consistencies between multimodal content and evidence items, facilitating multimodal OOC misinformation detection.

\begin{table}[t]
\caption{A multimodal OOC misinformation detectors with distinct LVLMs.}
\centering
        \begin{tabular}{lc|c|c}
            \toprule
            Setting & Tuning & Parameters & \textbf{All $\uparrow$} \\
            \hline
            Random &   & -  & 50.3	\\
            Qwen2-VL-2B &   & 1.5B  & 65.7 \\
            Qwen2-VL-7B &  & 7.6B  &  79.1 \\
            Qwen2-VL-72B &   & 72B  & 79.8 \\
            \hline
            Qwen2-VL-2B & \cmark & 1.5B  & 81.4	\\
            Qwen2-VL-7B & \cmark & 7.6B  & 89.9 \\
            \bottomrule
        \end{tabular}
\label{tab:5}
\end{table}

\noindent\textbf{Analysis of model size.} To assess the influence of different model sizes in the HiEAG, we start the investigation of HiEAG with Qwen2-VL family \cite{Qwen2VL} by varying model parameters from 2B to 72B in various experimental settings. As presented in Table \ref{tab:5}, we achieve the following findings: In the setting of zero-shot (see 2-th row and 4-th row in Table \ref{tab:5}), as model size increases, the detection performance across all samples improves. The results align with prior studies in scaling laws \cite{kaplan2020scalinglawsneurallanguage}. However, the base MLLM Qwen2-VL-72B just achieves a performance improvement of 0.7\%, and introduces around 10 times parameters, compared to Qwen2-VL-7B. This ignores the compromise between computation burden and detection performance. In the setting of instruction tuning, the general-purpose base MLLMs are extended to the task of OOC misinformation detection, resulting in significant improvements in performance. This encourages us to select Qwen2-VL-7B as the base MLLM.

\noindent\textbf{Analysis of training data size.} To evaluate the feasibility of rapid deployment of HiEAG at the stage of early detection, we vary the size of training data from 10\% to 75\% while fixing other configurations during the training process. As shown in Fig. \ref{fig:5}, our proposed method HiEAG achieves remarkable performance in any experimental settings. With the growth of training data, the discernment capabilities of HiEAG improves, resulting in a rising tendency in the term of accuracy across all samples. This highlights that HiEAG can be rapidly deployed while ensuring comparable detection accuracy, even at the early stage.

\begin{figure}[t]
  \centering
   \includegraphics[width=\linewidth]{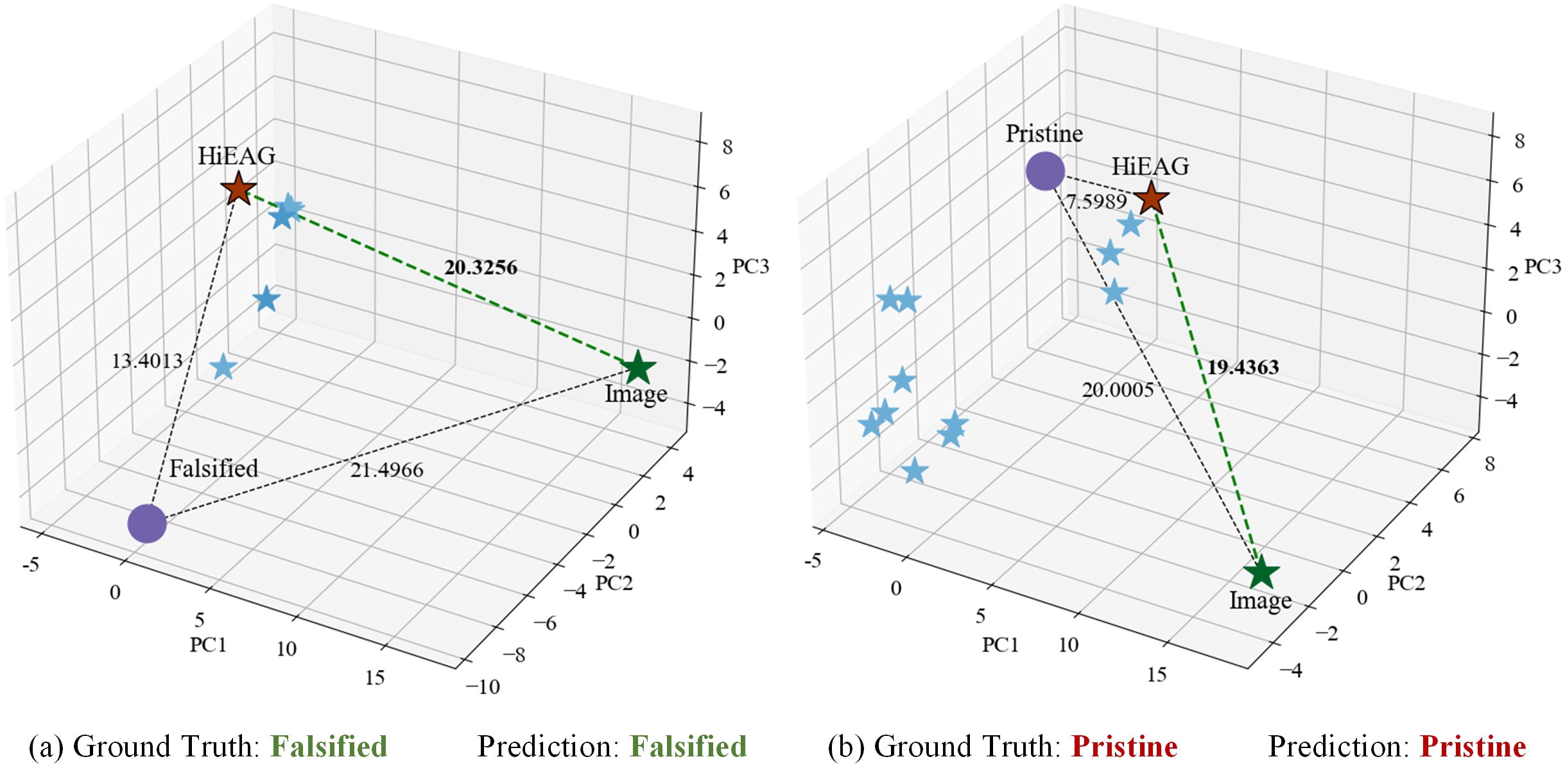}
   \caption{\textbf{Visualization of different data distributions.} Subfigure (a) represents a falsified image-text pair. Subfigure (b) represents a pristine image-text pair. Blue points denote the retrieved evidence regarding image-text pairs.}
   \label{fig:6}
\end{figure}

\subsection{Case study and visualization}

As presented in Fig. \ref{fig:6}, we provide two instances from the validation set of NewsCLIPpings \cite{luo-etal-2021-newsclippings} to demonstrate the effectiveness of HiEAG through visualization. The instances provide valuable insights regarding the contributions of HiEAG in the OOC misinformation detection task. In the first instance, by analyzing an OOC image-text pair (see subfigure (a) in Fig. \ref{fig:6}), we note that the novel generated sentence is far further away from the image-text pair. This prevents the model from modeling the relationships between the image and its misleading textual context. In the second instance, by observing a pristine image-text pair (see subfigure (b) in Fig. \ref{fig:6}), we find that HiEAG maintains the generated sentence close to the image-text pair, thereby facilitating an accurate judgment of the targeted content. The observation above demonstrates the effectiveness and superiority of our proposed HiEAG in refining consistency checking between multimodal content and external information, highlighting its capacity to the task of OOC misinformation detection.
\section{Conclusion}
\label{sec:Conclusion}

In this paper, we presented HiEAG, a novel hierarchical evidence-augmented generation framework designed to the task of multimodal out-of-context misinformation detection. In the evidence reranking, we designed Automatic Evidence Selection Prompting to learn the relevant evidence item, enhancing evidence retrieval for multimodal content. In the evidence rewriting, we devised Automatic Evidence Generation Prompting to achieve the alignment sentence, which improves task adaptation on model architectures. Additionally, to enable the model for both explanation and judgment, we based on the instruction data, extended instruction tuning to the task for training a detector. Through extensive experiments on synthetic and real-world datasets, we demonstrated the effectiveness and robustness of the proposed HiEAG in addressing the challenges of multimodal out-of-context misinformation detection.
{
    \small
    \bibliographystyle{ieeenat_fullname}
    \bibliography{main}

@inproceedings{luo-etal-2021-newsclippings,
    title = "{N}ews{CLIP}pings: {A}utomatic {G}eneration of {O}ut-of-{C}ontext {M}ultimodal {M}edia",
    author = "Luo, Grace  and
      Darrell, Trevor  and
      Rohrbach, Anna",
    booktitle = "Proceedings of 2021 Conference on Empirical Methods in Natural Language Processing (EMNLP)",
    year = "2021",
    pages = "6801--6817",
}

@inproceedings{li-etal-2025-cmie,
    title = "{CMIE}: Combining {MLLM} Insights with External Evidence for Explainable Out-of-Context Misinformation Detection",
    author = "Li, Fanxiao  and
      Wu, Jiaying  and
      He, Canyuan  and
      Zhou, Wei",
    booktitle = "Proceedings of the Findings of the Association for Computational Linguistics: ACL 2025",
    year = "2025",
    pages = "9342--9354",
}

@inproceedings{ma-etal-2024-event,
    title = "Event-Radar: Event-driven Multi-View Learning for Multimodal Fake News Detection",
    author = "Ma, Zihan  and
      Luo, Minnan  and
      Guo, Hao  and
      Zeng, Zhi  and
      Hao, Yiran  and
      Zhao, Xiang",
    booktitle = "Proceedings of the 62th Annual Meeting of the Association for Computational Linguistics (ACL)",
    year = "2024",
    pages = "5809--5821",
}

@article{nan2025exploiting,
  title={Exploiting user comments for early detection of fake news prior to users’ commenting},
  author={Nan, Qiong and Sheng, Qiang and Cao, Juan and Zhu, Yongchun and Wang, Danding and Yang, Guang and Li, Jintao},
  journal={Frontiers of Computer Science},
  volume={19},
  number={10},
  pages={1910354},
  year={2025},
  publisher={Springer}
}

@article{aslett2024online,
  title={Online searches to evaluate misinformation can increase its perceived veracity},
  author={Aslett, Kevin and Sanderson, Zeve and Godel, William and Persily, Nathaniel and Nagler, Jonathan and Tucker, Joshua A},
  journal={Nature},
  volume={625},
  number={7995},
  pages={548--556},
  year={2024},
  publisher={Nature Publishing Group UK London}
}

@article{lakara2025llmconsensusmultiagentdebatevisual,
  title={LLM-Consensus: Multi-Agent Debate for Visual Misinformation Detection},
  author={Lakara, Kumud and Channing, Georgia and Sock, Juil and Rupprecht, Christian and Torr, Philip and Collomosse, John and de Witt, Christian Schroeder},
  year={2024, arXiv preprint arXiv: 2410.20140},
}

@InProceedings{Qi_2024_CVPR,
    author    = {Qi, Peng and Yan, Zehong and Hsu, Wynne and Lee, Mong Li},
    title     = {SNIFFER: Multimodal Large Language Model for Explainable Out-of-Context Misinformation Detection},
    booktitle = {Proceedings of 2024 IEEE/CVF Conference on Computer Vision and Pattern Recognition (CVPR)},
    year      = {2024},
    pages     = {13052--13062}
}

@inproceedings{Xu_MMOOC,
    title = "MMOOC: A Multimodal Misinformation Dataset for Out-of-Context News Analysis",
    author="Xu, Qingzheng
    and Du, Heming
    and Chen, Huiqiang
    and Liu, Bo
    and Yu, Xin",
    booktitle = "Proceedings of the 29th Australasian Conference on Information Security and Privacy (ACISP)",
    year = "2024",
    pages = "444--459",
}

@inproceedings{Jin_2017,
author = {Jin, Zhiwei and Cao, Juan and Guo, Han and Zhang, Yongdong and Luo, Jiebo},
title = {Multimodal Fusion with Recurrent Neural Networks for Rumor Detection on Microblogs},
year = {2017},
booktitle = {Proceedings of the 25th ACM International Conference on Multimedia (MM)},
pages = {795--816}
}

@article{vector_Salton,
author = {Salton, G. and Wong, A. and Yang, C. S.},
title = {A vector space model for automatic indexing},
year = {1975},
volume = {18},
number = {11},
journal={Communications of the ACM},
pages = {613--620},
numpages = {8},
}

@article{Qwen2VL,
  title={Qwen2-VL: Enhancing Vision-Language Model's Perception of the World at Any Resolution},
  author={Wang, Peng and Bai, Shuai and Tan, Sinan and Wang, Shijie and Fan, Zhihao and Bai, Jinze and Chen, Keqin and Liu, Xuejing and Wang, Jialin and Ge, Wenbin and Fan, Yang and Dang, Kai and Du, Mengfei and Ren, Xuancheng and Men, Rui and Liu, Dayiheng and Zhou, Chang and Zhou, Jingren and Lin, Junyang},
  journal={2024, arXiv preprint arXiv: 2409.12191},
}

@inproceedings{Papadopoulos_DT,
author = {Papadopoulos, Stefanos-Iordanis and Koutlis, Christos and Papadopoulos, Symeon and Petrantonakis, Panagiotis},
title = {Synthetic Misinformers: Generating and Combating Multimodal Misinformation},
year = {2023},
booktitle = {Proceedings of the 2nd ACM International Workshop on Multimedia AI against Disinformation (MAD)},
pages = {36–44},
numpages = {9}
}

@inproceedings{Aneja_Bregler_Niessner_2023,
    title = "COSMOS: Catching Out-of-Context Image Misuse Using Self-Supervised Learning",
    author = "Aneja, Shivangi and Bregler, Chris and Niessner, Matthias",
    booktitle = "Proceedings of the 39th AAAI Conference on Artificial Intelligence (AAAI)",
    year = "2025",
    pages = "9342--9354",
}

@inproceedings{Jaiswal_MAIM,
author = {Jaiswal, Ayush and Sabir, Ekraam and AbdAlmageed, Wael and Natarajan, Premkumar},
title = {Multimedia Semantic Integrity Assessment Using Joint Embedding Of Images And Text},
year = {2017},
isbn = {9781450349062},
booktitle = {Proceedings of the 25th ACM International Conference on Multimedia (MM)},
pages = {1465--1471},
}

@inproceedings{Sabir_MEIR,
author = {Sabir, Ekraam and AbdAlmageed, Wael and Wu, Yue and Natarajan, Prem},
title = {Deep Multimodal Image-Repurposing Detection},
year = {2018},
booktitle = {Proceedings of the 26th ACM International Conference on Multimedia (MM)},
pages = {1337–1345},
}

@InProceedings{Jaiswal_2019_CVPR,
author = {Jaiswal, Ayush and Wu, Yue and AbdAlmageed, Wael and Masi, Iacopo and Natarajan, Premkumar},
title = {AIRD: Adversarial Learning Framework for Image Repurposing Detection},
booktitle = {Proceedings of 2019 IEEE/CVF Conference on Computer Vision and Pattern Recognition (CVPR)},
year = {2019},
pages = {11330-11339}
}

@inproceedings{Budack_TamperedNews,
author = {M\"{u}ller-Budack, Eric and Theiner, Jonas and Diering, Sebastian and Idahl, Maximilian and Ewerth, Ralph},
title = {Multimodal Analytics for Real-world News using Measures of Cross-modal Entity Consistency},
year = {2020},
booktitle = {Proceedings of 2020 International Conference on Multimedia Retrieval (ICMR)},
pages = {16–25},
}

@inproceedings{zlatkova-etal-2019-fact,
    title = "Fact-Checking Meets Fauxtography: Verifying Claims About Images",
    author = "Zlatkova, Dimitrina  and
      Nakov, Preslav  and
      Koychev, Ivan",
    booktitle = "Proceedings of 2019 Conference on Empirical Methods in Natural Language Processing and the 9th International Joint Conference on Natural Language Processing (EMNLP-IJCNLP)",
    year = "2019",
    pages = "2099--2108",
}

@InProceedings{Abdelnabi_2022_CVPR,
    author    = {Abdelnabi, Sahar and Hasan, Rakibul and Fritz, Mario},
    title     = {Open-Domain, Content-Based, Multi-Modal Fact-Checking of Out-of-Context Images via Online Resources},
    booktitle = {Proceedings of 2022 IEEE/CVF Conference on Computer Vision and Pattern Recognition (CVPR)},
    year      = {2022},
    pages     = {14940-14949}
}

@inproceedings{yuan-etal-2023-support,
    title = "Support or Refute: Analyzing the Stance of Evidence to Detect Out-of-Context Mis- and Disinformation",
    author = "Yuan, Xin  and
      Guo, Jie  and
      Qiu, Weidong  and
      Huang, Zheng  and
      Li, Shujun",
    booktitle = "Proceedings of 2023 Conference on Empirical Methods in Natural Language Processing (EMNLP)",
    year = "2023",
    pages = "4268--4280",
}

@InProceedings{pmlr-v139-radford21a,
  title = 	 {Learning Transferable Visual Models From Natural Language Supervision},
  author =       {Radford, Alec and Kim, Jong Wook and Hallacy, Chris and Ramesh, Aditya and Goh, Gabriel and Agarwal, Sandhini and Sastry, Girish and Askell, Amanda and Mishkin, Pamela and Clark, Jack and Krueger, Gretchen and Sutskever, Ilya},
  booktitle = 	 {Proceedings of the 38th International Conference on Machine Learning (ICML)},
  pages = 	 {8748--8763},
  year = 	 {2021},
  volume = 	 {139}
}

@inproceedings{Dai_InstructBLIP,
author = {Dai, Wenliang and Li, Junnan and Li, Dongxu and Tiong, Anthony Meng Huat and Zhao, Junqi and Wang, Weisheng and Li, Boyang and Fung, Pascale and Hoi, Steven},
title = {InstructBLIP: towards general-purpose vision-language models with instruction tuning},
year = {2023},
publisher = {Curran Associates Inc.},
address = {Red Hook, NY, USA},
booktitle = {Proceedings of the 37th International Conference on Neural Information Processing Systems},
articleno = {2142},
numpages = {18},
location = {New Orleans, LA, USA},
series = {NIPS '23}
}

@misc{vicuna2023,
    title = {Vicuna: An Open-Source Chatbot Impressing GPT-4 with 90\%* ChatGPT Quality},
    author = {Chiang, Wei-Lin and Li, Zhuohan and Lin, Zi and Sheng, Ying and Wu, Zhanghao and Zhang, Hao and Zheng, Lianmin and Zhuang, Siyuan and Zhuang, Yonghao and Gonzalez, Joseph E. and Stoica, Ion and Xing, Eric P.},
    year = {2023}
}

@article{gu2025multimllmknowledgedistillationoutofcontext,
  title={Multi-MLLM Knowledge Distillation for Out-of-Context News Detection},
  author={Yimeng Gu and Zhao Tong and Ignacio Castro and Shu Wu and Gareth Tyson},
  journal={2025, arXiv preprint arXiv: 2505.22517},
}

@InProceedings{Jiang_2024_CVPR,
    author    = {Jiang, Chaoya and Xu, Haiyang and Dong, Mengfan and Chen, Jiaxing and Ye, Wei and Yan, Ming and Ye, Qinghao and Zhang, Ji and Huang, Fei and Zhang, Shikun},
    title     = {Hallucination Augmented Contrastive Learning for Multimodal Large Language Model},
    booktitle = {Proceedings of 2024 IEEE/CVF Conference on Computer Vision and Pattern Recognition (CVPR)},
    year      = {2024},
    pages     = {27036-27046}
}

@InProceedings{Liu_2024_CVPR,
    author    = {Liu, Huan and Tan, Zichang and Tan, Chuangchuang and Wei, Yunchao and Wang, Jingdong and Zhao, Yao},
    title     = {Forgery-aware Adaptive Transformer for Generalizable Synthetic Image Detection},
    booktitle = {Proceedings of 2024 IEEE/CVF Conference on Computer Vision and Pattern Recognition (CVPR)},
    year      = {2024},
    pages     = {10770-10780}
}

@article{Liu_UFAFormer,
author = {Liu, Huan and Tan, Zichang and Chen, Qiang and Wei, Yunchao and Zhao, Yao and Wang, Jingdong},
title = {Unified Frequency-Assisted Transformer Framework for Detecting and Grounding Multi-modal Manipulation},
year = {2024},
volume = {133},
number = {3},
journal = {International Journal of Computer Vision},
pages = {1392–1409},
numpages = {18},
}

@inproceedings{Liu_LLaVA,
author = {Liu, Haotian and Li, Chunyuan and Wu, Qingyang and Lee, Yong Jae},
title = {Visual instruction tuning},
year = {2023},
booktitle = {Proceedings of the 37th International Conference on Neural Information Processing Systems (NIPS)},
articleno = {1516},
pages = {34892--34916},
}

@article{chen2023minigptv2largelanguagemodel,
  title={MiniGPT-v2: large language model as a unified interface for vision-language multi-task learning},
  author={Jun Chen and Deyao Zhu and Xiaoqian Shen and Xiang Li and Zechun Liu and Pengchuan Zhang and Raghuraman Krishnamoorthi and Vikas Chandra and Yunyang Xiong and Mohamed Elhoseiny},
  journal={2023 arXiv preprint arXiv: 2310.09478},
}

@InProceedings{Liu_LLaVA15,
    author    = {Liu, Haotian and Li, Chunyuan and Li, Yuheng and Lee, Yong Jae},
    title     = {Improved Baselines with Visual Instruction Tuning},
    booktitle = {Proceedings of 2024 IEEE/CVF Conference on Computer Vision and Pattern Recognition (CVPR)},
    year      = {2024},
    pages     = {26296-26306}
}

@inproceedings{Han_ZSVQA,
author = {Han, Seunghoon and Choi, Mingyu and Lee, Hyewon and Park, SoYoung and Lee, Jong-Ryul and Lim, Sungsu and Kim, Tae-Ho},
title = {Diverse Knowledge Selection for Enhanced Zero-shot Visual Question Answering},
year = {2025},
booktitle = {Proceedings of the ACM on Web Conference 2025 (WWW)},
pages = {2161–2169},
}

@article{YUAN2025111783,
  title={Exp-VQA: fine-grained facial expression analysis via visual question answering},
  author={Yuan, Yujian and Zeng, Jiabei and Shan, Shiguang},
volume = {168},
  journal={Pattern Recognition},
  pages={111783},
  year={2025}
}

@article{LI2025129122,
title = {ClipSAM: CLIP and SAM collaboration for zero-shot anomaly segmentation},
journal = {Neurocomputing},
volume = {618},
pages = {129122},
year = {2025},
author = {Shengze Li and Jianjian Cao and Peng Ye and Yuhan Ding and Chongjun Tu and Tao Chen},
}

@article{PENG2025113176,
title = {SAM-LAD: Segment Anything Model meets zero-shot logic anomaly detection},
journal = {Knowledge-Based Systems},
volume = {314},
pages = {113176},
year = {2025},
issn = {0950-7051},
doi = {https://doi.org/10.1016/j.knosys.2025.113176},
url = {https://www.sciencedirect.com/science/article/pii/S0950705125002230},
author = {Yun Peng and Xiao Lin and Nachuan Ma and Jiayuan Du and Chuangwei Liu and Chengju Liu and Qijun Chen}
}

@article{zhou2024correctingmisinformationsocialmedia,
  title={Correcting misinformation on social media with a large language model},
  author={Xinyi Zhou and Ashish Sharma and Amy X. Zhang and Tim Althoff},
  year={2024, arXiv preprint arXiv: 2403.11169},
}

@article{Xu_ZSVQA,
author = {Xu, Liyong and Jiao, Yifan and Bao, Bing-Kun},
title = {Bool Prompt with Decomposition and Enhancement: Zero-Shot VQA Based on PVLMs},
year = {2025},
volume = {21},
number = {9},
pages = {1--21},
journal = {ACM Transactions on Multimedia Computing, Communications and Applications},
}

@inproceedings{Xu_Sun_Zhai_Li_Liang_Li_Du_2025,
author = {Xu, Yifang and Sun, Yunzhuo and Zhai, Benxiang and Li, Ming and Liang, Wenxin and Li, Yang and Du, Sidan},
title = {Zero-shot Video Moment Retrieval via Off-the-shelf Multimodal Large Language Models},
year = {2025},
booktitle = {Proceedings of the AAAI Conference on Artificial Intelligence (AAAI)},
pages = {8978-8986},
}

@article{papadopoulos2024verite,
  title={VERITE: a Robust benchmark for multimodal misinformation detection accounting for unimodal bias},
  author={Papadopoulos, Stefanos-Iordanis and Koutlis, Christos and Papadopoulos, Symeon and Petrantonakis, Panagiotis C},
  journal={International Journal of Multimedia Information Retrieval},
  volume={13},
  number={1},
  pages={4},
  year={2024},
}

@inproceedings{liu-etal-2021-visual,
    title = "Visual News: Benchmark and Challenges in News Image Captioning",
    author = "Liu, Fuxiao  and
      Wang, Yinghan  and
      Wang, Tianlu  and
      Ordonez, Vicente",
    booktitle = "Proceedings of 2021 Conference on Empirical Methods in Natural Language Processing (EMNLP)",
    year = "2021",
    pages = "6761--6771",
}

@article{wu2025exclaimexplainablecrossmodalagentic,
  title={EXCLAIM: An Explainable Cross-Modal Agentic System for Misinformation Detection with Hierarchical Retrieval},
  author={Yin Wu and Zhengxuan Zhang and Fuling Wang and Yuyu Luo and Hui Xiong and Nan Tang},
  year={2025, arXiv preprint arXiv: 2504.06269},
}

@article{xiao2025xfacta,
  title={XFacta: Contemporary, Real-World Dataset and Evaluation for Multimodal Misinformation Detection with Multimodal LLMs},
  author={Xiao, Yuzhuo and Han, Zeyu and Wang, Yuhan and Jiang, Huaizu},
  journal={2025, arXiv preprint arXiv: 2508.09999},
}

@article{yan2025mitigatinggenaipoweredevidencepollution,
  title={Mitigating GenAI-powered Evidence Pollution for Out-of-Context Multimodal Misinformation Detection},
  author={Zehong Yan and Peng Qi and Wynne Hsu and Mong Li Lee},
  year={2025, arXiv preprint arXiv: 2501.14728},
}

@article{xuan2024lemmalvlmenhancedmultimodalmisinformation,
  title={LEMMA: Towards LVLM-Enhanced Multimodal Misinformation Detection with External Knowledge Augmentation},
  author={Keyang Xuan and Li Yi and Fan Yang and Ruochen Wu and Yi R. Fung and Heng Ji},
  journal={2024, arXiv preprint arXiv: 2402.11943},
}

@InProceedings{10.1007/978-981-96-8180-8_25,
author="Yang, Dayu
and Wang, Fuli",
title="Towards Retrieval-Augmented Large Language Model-Based Conversational Recommender System",
booktitle="Proceedings of the 29th Pacific-Asia Conference on Knowledge Discovery and Data Mining (PAKDD)",
year="2025",
pages="317--330",
}

@article{10.1145/3701228,
author = {Lyu, Yuanjie and Li, Zhiyu and Niu, Simin and Xiong, Feiyu and Tang, Bo and Wang, Wenjin and Wu, Hao and Liu, Huanyong and Xu, Tong and Chen, Enhong},
title = {CRUD-RAG: A Comprehensive Chinese Benchmark for Retrieval-Augmented Generation of Large Language Models},
year = {2025},
volume = {43},
number = {2},
journal = {ACM Transactions on Information Systems},
pages="1--32",
}

@inproceedings{malviya-katsigiannis-2024-evidence,
    title = "Evidence Retrieval for Fact Verification using Multi-stage Reranking",
    author = "Malviya, Shrikant  and
      Katsigiannis, Stamos",
    booktitle = "Proceedings of the Findings of the Association for Computational Linguistics: EMNLP 2024",
    year = "2024",
    pages = "7295--7308",
}

@InProceedings{10.1007/978-3-030-47436-2_27,
author="Zhou, Xinyi
and Wu, Jindi
and Zafarani, Reza",
title="SAFE: Similarity-Aware Multi-modal Fake News Detection",
booktitle="Proceedings of the 24th Pacific-Asia Conference on Knowledge Discovery and Data Mining (PAKDD)",
year="2020",
pages="354--367",
}

@inproceedings{10.1145/3219819.3219903,
author = {Wang, Yaqing and Ma, Fenglong and Jin, Zhiwei and Yuan, Ye and Xun, Guangxu and Jha, Kishlay and Su, Lu and Gao, Jing},
title = {EANN: Event Adversarial Neural Networks for Multi-Modal Fake News Detection},
year = {2018},
booktitle = {Proceedings of the 24th ACM SIGKDD International Conference on Knowledge Discovery and Data Mining (KDD)},
pages = {849–857},
}

@article{li2019visualbertsimpleperformantbaseline,
  title={VisualBERT: A Simple and Performant Baseline for Vision and Language},
  author={Liunian Harold Li and Mark Yatskar and Da Yin and Cho-Jui Hsieh and Kai-Wei Chang},
  year={2019, arXiv preprint arXiv: 1908.03557},
}

@article{zhang2024interpretabledetectionoutofcontextmisinformation,
  title={Interpretable Detection of Out-of-Context Misinformation with Neural-Symbolic-Enhanced Large Multimodal Model},
  author={Yizhou Zhang and Loc Trinh and Defu Cao and Zijun Cui and Yan Liu},
  year={2024, arXiv preprint arXiv: 2304.07633},
}

@InProceedings{10.1007/978-3-031-57916-5_8,
author="Shalabi, Fatma
and Felouat, Hichem
and Nguyen, Huy H.
and Echizen, Isao",
title="Leveraging Chat-Based Large Vision Language Models for Multimodal Out-of-Context Detection",
booktitle="Proceedings of the 38th International Conference on Advanced Information Networking and Applications (AINA)",
year="2024",
pages="86--98",
}

@InProceedings{10.5555/3454287.3455008,
author = {Paszke, Adam and Gross, Sam and Massa, Francisco and Lerer, Adam and Bradbury, James and Chanan, Gregory and Killeen, Trevor and Lin, Zeming and Gimelshein, Natalia and Antiga, Luca and Desmaison, Alban and K\"{o}pf, Andreas and Yang, Edward and DeVito, Zach and Raison, Martin and Tejani, Alykhan and Chilamkurthy, Sasank and Steiner, Benoit and Fang, Lu and Bai, Junjie and Chintala, Soumith},
title = {PyTorch: an imperative style, high-performance deep learning library},
year = {2019},
booktitle = {Proceedings of the 33rd International Conference on Neural Information Processing Systems (NIPS)},
pages={8026-8037}
}

@article{loshchilov2019decoupledweightdecayregularization,
    author = "Ilya Loshchilov and Frank Hutter",
    title = "Decoupled Weight Decay Regularization",
    journal = "2019, arXiv preprint arXiv: 1711.05101",
}

@inproceedings{su-etal-2023-pandagpt,
    title = "{P}anda{GPT}: One Model To Instruction-Follow Them All",
    author = "Su, Yixuan  and
      Lan, Tian  and
      Li, Huayang  and
      Xu, Jialu  and
      Wang, Yan  and
      Cai, Deng",
    booktitle = "Proceedings of the 1st Workshop on Taming Large Language Models (TLLM)",
    year = "2023",
    pages = "11--23",
}

@inproceedings{dao2023flashattention2,
  title={Flash{A}ttention-2: Faster Attention with Better Parallelism and Work Partitioning},
  author={Dao, Tri},
  booktitle={Proceedings of the 12th International Conference on Learning Representations (ICLR)},
  year={2024}
}

@article{kaplan2020scalinglawsneurallanguage,
  title={Scaling Laws for Neural Language Models},
  author={Jared Kaplan and Sam McCandlish and Tom Henighan and Tom B. Brown and Benjamin Chess and Rewon Child and Scott Gray and Alec Radford and Jeffrey Wu and Dario Amodei},
  journal={2020, arXiv preprint arXiv: 2001.08361},
}

@article{zhu2023minigpt4enhancingvisionlanguageunderstanding,
  title={MiniGPT-4: Enhancing Vision-Language Understanding with Advanced Large Language Models},
  author={Deyao Zhu and Jun Chen and Xiaoqian Shen and Xiang Li and Mohamed Elhoseiny},
  year={2023, arXiv preprint arXiv: 2304.10592},
}

@inproceedings{hu2022lora,
  title={Lo{RA}: Low-Rank Adaptation of Large Language Models},
  author={Edward J Hu and Yelong Shen and Phillip Wallis and Zeyuan Allen-Zhu and Yuanzhi Li and Shean Wang and Lu Wang and Weizhu Chen},
    booktitle={Proceedings of the 10th International Conference on Learning Representations (ICLR)},
     year={2022}
}
}


\end{document}